\documentclass[conference]{IEEEtran}
\IEEEoverridecommandlockouts

\usepackage{cite}
\usepackage{amsmath,amssymb,amsfonts}
\usepackage{algorithmic}
\usepackage{graphicx}
\usepackage{textcomp}

\usepackage{hyperref}
\usepackage[table,xcdraw]{xcolor} % tables
\usepackage{multirow} % tables merge cells

\usepackage[normalem]{ulem}

\def\BibTeX{{\rm B\kern-.05em{\sc i\kern-.025em b}\kern-.08em
    T\kern-.1667em\lower.7ex\hbox{E}\kern-.125emX}}
\begin{document}

\title{An Island-Based Parallel Biased Random-Key Genetic Algorithm for the Three-Dimensional Trailer Loading Problem}

\author{
\IEEEauthorblockN{
Alfredo del Río Moldes,
Lucía Díaz Rodríguez,
Luis Carlos de Vicente Poutás,\\
Javier Cameselle Abreu,
Bruno Fernández Castro
}
\IEEEauthorblockA{
Intelligent Systems Department\\
Gradiant\\
Vigo, Spain\\
\{adelrio, ldiaz, cdevicente, jcameselle, bfernandez\}@gradiant.org
}
}

\maketitle
\thispagestyle{plain}
\pagestyle{plain}

\begin{abstract}
The Three-Dimensional Trailer Loading Problem (3D-TLP) involves determining the optimal placement and orientation of heterogeneous items within the confined space of a trailer while maximizing volume utilization and satisfying a wide range of complex logistical and safety constraints. The 3D-TLP is NP-hard, rendering exact optimization approaches computationally impractical for large-scale industrial applications. To address this challenge, we propose an enhanced Biased Random-Key Genetic Algorithm (BRKGA) accelerated through a novel island-based parallelization framework, PANGEA. The proposed method combines the search efficiency and robustness of BRKGA with a multi-population evolutionary scheme for genetic algorithms. This island-model strategy promotes population diversity, mitigates premature convergence, and significantly reduces computational times. The proposed solution was validated in a real trailer loading process, providing an effective solution approach for real-world large-scale logistics. 
\end{abstract}

\begin{IEEEkeywords}
biased random-key genetic algorithm, island-based genetic algorithm, empty maximal spaces, three-dimensional trailer loading problem, logistics optimization, load planning
\end{IEEEkeywords}

\section{INTRODUCTION}
The efficient loading of goods into trailers is a fundamental problem in modern logistics and supply chain management. As transportation costs continue to represent a significant portion of total logistics expenses, maximizing the utilization of available cargo space while satisfying operational and safety constraints has become an increasingly important objective. The Three-Dimensional Trailer Loading Problem (3D-TLP) consists of determining the optimal placement and orientation of a set of heterogeneous items within the limited volume of a trailer, seeking to maximize space utilization while complying with a variety of logistic and security constraints.

The 3D-TLP belongs to the class of three-dimensional packing problems and is known to be NP-hard, making the computation of exact optimal solutions computationally intractable for large-scale industrial instances. Beyond the inherent combinatorial complexity, real-world trailer loading involves numerous additional requirements that substantially increase the difficulty of the problem. These include weight distribution to ensure vehicle stability, axle load limitations imposed by transportation regulations, cargo compatibility, load-bearing capacities, unloading sequence constraints, and the prevention of cargo movement during transportation. Improper load arrangements may compromise vehicle handling, increase fuel consumption, damage transported goods, or even pose serious road safety risks.

In practical logistics environments, solution quality is only one aspect of the problem. Transportation planning is typically performed under strict operational deadlines, where loading plans must be generated within minutes to support warehouse operations and dispatch schedules. Consequently, optimization methods must not only produce high-quality loading configurations but also satisfy stringent computational time requirements. This need for fast and robust decision-making has motivated the development of heuristic and metaheuristic approaches capable of exploring large search spaces efficiently.

Among these approaches, Genetic Algorithms (GAs) have proven particularly suitable for solving complex combinatorial optimization problems due to their ability to balance exploration and exploitation while maintaining a diverse population of candidate solutions. Inspired by the principles of natural evolution, GAs iteratively improve solution quality through selection, crossover, and mutation operators. Their population-based nature makes them especially attractive for highly constrained packing problems, where multiple promising regions of the search space can be explored simultaneously.

A particularly successful variant is the Biased Random-Key Genetic Algorithm (BRKGA), which represents solutions as vectors of random keys that are decoded into feasible loading configurations. This representation simplifies the design of genetic operators, preserves solution feasibility through specialized decoding procedures, and has demonstrated remarkable performance across a wide range of combinatorial optimization problems, including packing, routing, scheduling, and resource allocation. Furthermore, the independence between individuals naturally facilitates parallelization, allowing different populations to evolve concurrently while periodically exchanging high-quality solutions.

This natural parallel structure finds a compelling analogy in biological evolution. In nature, geographically isolated populations accumulate distinct adaptations over time, each following its own evolutionary trajectory shaped by local pressures. When these populations eventually come into contact, the exchange of genetic material introduces novel combinations that neither lineage could have produced in isolation, triggering bursts of evolutionary innovation and preventing the genetic stagnation that results from prolonged inbreeding. The island model of parallel genetic algorithms mirrors this process directly: each subpopulation explores a different region of the search space independently, avoiding premature convergence toward a single local optimum, while periodic migration of elite individuals plays the role of inter-population genetic exchange --- injecting diversity, disrupting stagnant trajectories, and accelerating the discovery of high-quality solutions. The structure of the migration network, analogous to the geography that governs how isolated populations come into contact, further determines the speed and reach of this information flow across the system.

 This work proposes a modified Biased Random-Key Genetic Algorithm accelerated through our novel island-based parallelization framework (PANGEA) for the Three-Dimensional Trailer Loading Problem. The proposed approach combines the robustness of BRKGAs with an island model parallelization strategy, where multiple subpopulations evolve independently to promote solution diversity and reduce the risk of premature convergence. Periodic migration of elite individuals enables information exchange between islands, improving both solution quality and convergence speed while taking advantage of modern multi-core computing architectures. Beyond the parallelization scheme itself, this work addresses two practical challenges that are often overlooked in the literature: how to fairly compare parallel configurations under identical computational budgets, and how the choice of migration topology affects convergence behavior in the context of a highly constrained packing problem.

The remainder of this paper is organized as follows. Section II reviews different state-of-the-art approaches aimed at solving the three-Dimensional Trailer Loading Problem. Section III introduces our distributed algorithm based on parallel island models for the biased random key genetic algorithm. Section IV describes the validation process, including metrics, environments, and results. Finally, Section V ends with conclusions and future work. 

\section{RELATED WORK}

The field of cutting and packing (C\&P) problems is fundamentally anchored in the efficient arrangement of items into larger objects, a challenge categorized by the improved typology of Wäscher et al. \cite{wascher_improved_2007} as either input minimization or output maximization \cite{alvarez-valdes_grasppath_2013, bortfeldt_constraints_2013}. The Bin Packing Problem (BPP) is a cornerstone of this domain, requiring the allocation of a strongly heterogeneous set of items into a minimum number of identical bins, known as the Single Bin-Size Bin Packing Problem (SBSBPP), or into bins of varying sizes and costs, termed the Multiple Bin-Size Bin Packing Problem (MBSBPP) \cite{bortfeldt_constraints_2013, alvarez-valdes_grasppath_2013, crainic_ts2pack_2009}. As these problems are recognized as NP-hard in the strong sense, especially in three-dimensional (3D) environments, research has increasingly moved from exact methods toward metaheuristic approaches to solve industrial-scale instances \cite{alvarez-valdes_grasppath_2013, lodi_heuristic_2002, lodi_tspack_2004}.

The transition from theoretical BPP to the practical 3D Trailer Loading Problem (3D-TLP) involves the integration of complex operational constraints that define the feasibility of a packing plan in real-world logistics \cite{alvarez-valdes_grasppath_2013, bortfeldt_constraints_2013, erbayrak_multi-objective_2021}. According to the comprehensive review by Bortfeldt and Wäscher \cite{bortfeldt_constraints_2013}, these constraints are categorized into item-related, container-related, and load-related factors, including orientation restrictions where items must maintain a "this side up" position while potentially allowing 90-degree horizontal rotations \cite{abeysooriya_jostle_2018, bortfeldt_constraints_2013, goncalves_parallel}. Stability is a critical concern, subdivided into vertical (static) stability, ensuring items are supported from below to prevent falling, and horizontal (dynamic) stability to prevent shifting during transit \cite{bortfeldt_parallel_2003, bortfeldt_constraints_2013, erbayrak_multi-objective_2021, pollaris_vehicle_2015}. Furthermore, industrial trailer loading must account for weight distribution and axle weight limits, as unbalanced loads can lead to vehicle instability, road erosion, and severe legal penalties \cite{bortfeldt_constraints_2013, krebs_axle_2021, pollaris_loading_2018}. Advanced loading scenarios also incorporate multi-drop constraints, necessitating a Last-In-First-Out (LIFO) unloading sequence to ensure that items for early destinations are not blocked by those intended for later stops \cite{bortfeldt_constraints_2013, erbayrak_multi-objective_2021, pollaris_vehicle_2015, pollaris_loading_2018}.

To address the geometric complexity of 3D-TLP, researchers have developed various spatial representation methods for tracking available volume within a bin \cite{crainic_ts2pack_2009, goncalves_parallel, zhao_comparative_2016}. The Corner Point (CP) method defines candidate positions where an item can be placed without being dominated by other potential locations, effectively reducing the search space \cite{crainic_ts2pack_2009, harrath_three-stage_2022, martello_three-dimensional_2000}. Alternatively, the Empty Maximal Space (EMS) representation identifies the largest empty cuboids available, allowing for a more flexible and dense arrangement of boxes \cite{goncalves_parallel, goncalves_biased_2013, zhao_comparative_2016, alvarez-valdes_grasppath_2013}. For scenarios involving irregular shapes or non-convex items, more sophisticated tools like the No-Fit Polygon (NFP) and Inner Fit Polygon (IFP) are employed to prevent overlaps while allowing for free or restricted rotation \cite{abeysooriya_jostle_2018, bortfeldt_constraints_2013, martinez-sykora_matheuristics_2017}.

The Biased Random Key Genetic Algorithm (BRKGA) has emerged as a particularly effective metaheuristic for solving these high-dimensional packing problems \cite{goncalves_parallel, goncalves_biased_2013, abeysooriya_jostle_2018, beltrao_brkga_2026}. BRKGA represents solutions as vectors of real numbers in the interval, which are then transformed into feasible packing sequences and orientations by a problem-specific decoder \cite{goncalves_parallel, goncalves_biased_2013, beltrao_brkga_2026, mandal_biased_2024}. A distinguishing feature of BRKGA is its "biased" crossover operator, where offspring have a higher probability (typically 70\%) of inheriting alleles from the elite population, ensuring the preservation of high-quality traits while mutants maintain population diversity \cite{goncalves_parallel, goncalves_biased_2013, beltrao_brkga_2026, mandal_biased_2024}. Recent advancements in BRKGA include the implementation of parallel multi-populations to prevent premature convergence and the development of specialized fitness functions that prioritize the emptying of the least-filled bins, which facilitates the reduction of the total bin count \cite{goncalves_parallel, goncalves_biased_2013, abeysooriya_jostle_2018, chu_parallel_nodate}.

Hybridization has further enhanced the performance of BRKGA in 3D bin packing and trailer loading \cite{alvarez-valdes_grasppath_2013, beltrao_brkga_2026, souto_q-learning_2024}. Zudio et al. \cite{zudio_brkgavnd_2018} combined BRKGA with Variable Neighborhood Descent (VND) to accelerate the search process by using "universal individuals" (pre-sorted sequences based on volume or dimensions) to seed the initial population. Similarly, Souto et al. \cite{souto_q-learning_2024} introduced a hybrid approach using Q-Learning to dynamically decide when to apply local search procedures based on the current solution gap, improving both computational efficiency and solution quality. In the context of irregular items, Junior et al. \cite{junior_biased_2020} successfully integrated BRKGA with a dotted board model to discretize irregular surfaces, enabling the handling of complex geometries in real-time packing applications.

Comparing BRKGA to other metaheuristics like Tabu Search (TS) and Guided Local Search (GLS) reveals its robustness across various benchmark sets \cite{alvarez-valdes_grasppath_2013, crainic_ts2pack_2009, zudio_brkgavnd_2018}. While unified tabu search codes such as TSpack provide a versatile framework for multi-dimensional bin packing, BRKGA often achieves superior volume utilization by evolving both the packing order and the choice of placement heuristics simultaneously \cite{lodi_tspack_2004, abeysooriya_jostle_2018, alvarez-valdes_grasppath_2013, crainic_ts2pack_2009}. The integration of 3D loading into routing problems, such as the 3L-CVRP, represents the current state of the art, where BRKGA-based decoders must satisfy all geometric, stability, and axle weight constraints while optimizing delivery paths \cite{pollaris_loading_2018, pollaris_vehicle_2015, bortfeldt_constraints_2013}. Despite the increasing complexity of these problems, the evolution from basic heuristic rules to intelligent, self-adaptive metaheuristics like the BRKGA continues to drive significant improvements in logistical efficiency and space utilization \cite{alvarez-valdes_grasppath_2013}.

Parallel Genetic Algorithms (PGAs) represent a sophisticated class of metaheuristics that achieve superior efficiency and efficacy compared to sequential models by utilizing structured populations \cite{alba_survey, cantu_survey}. Unlike basic sequential algorithms, PGAs can be implemented on various parallel architectures to simultaneously reduce search time and improve solution quality through localized competition and selection \cite{alba_survey, alba_parallel_2013, cantu_survey}. These algorithms are primarily classified into four structural models: global parallelization (master-slave), island models (coarse-grained distributed), cellular models (fine-grained), and hybrid paradigms \cite{alba_parallel_2013, alba_survey, harada_survey}. While the global model maintains a single panmictic population and parallelizes the computationally expensive fitness evaluation task, the other models fundamentally alter the search behavior by restricting individual interactions \cite{cantu_survey, harada_survey, mussi_evaluation_2011}.

\begin{figure*}[!ht]
\centering
\includegraphics[width=0.75\textwidth]{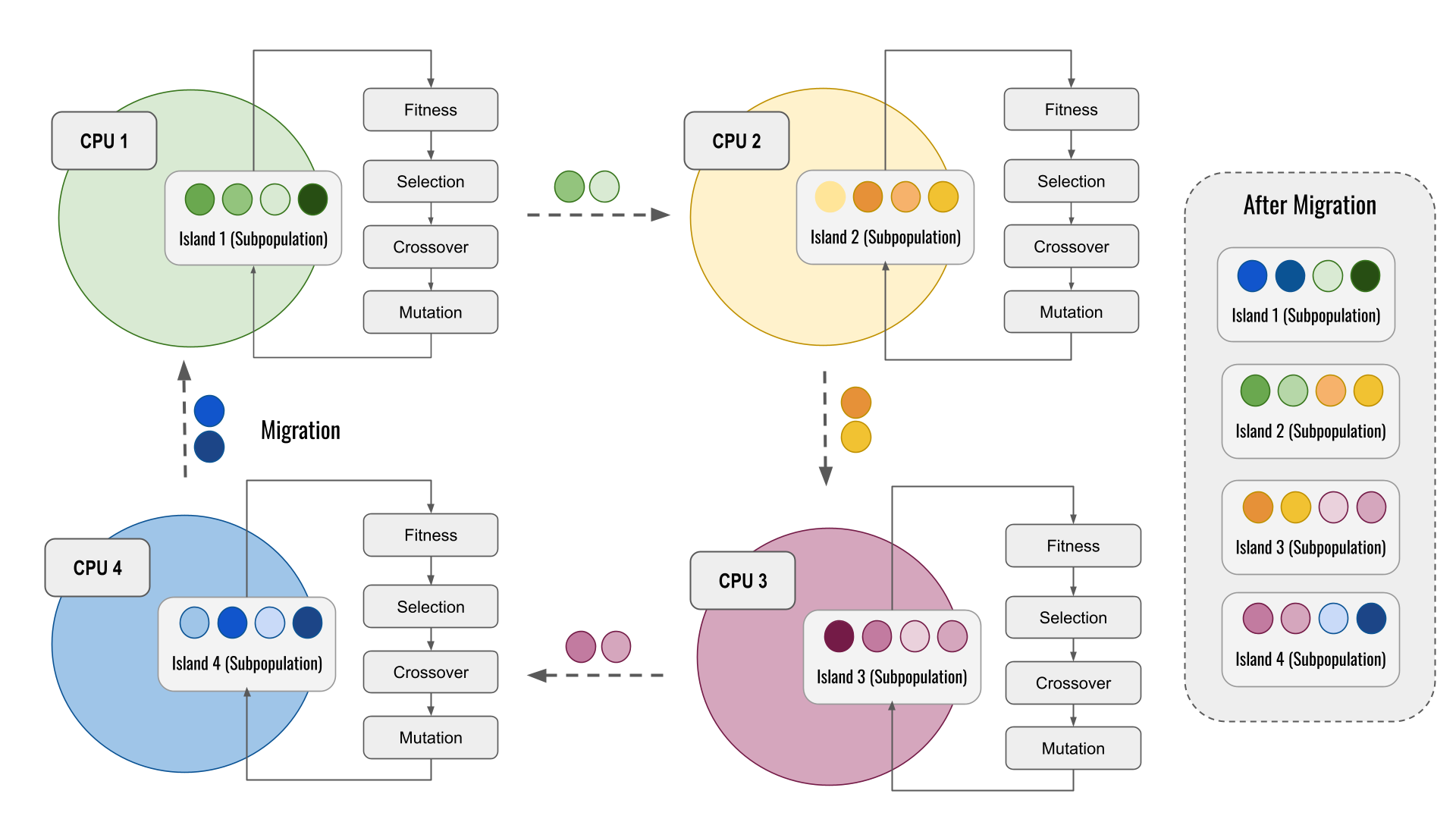}
\caption{Parallel Island Model. Source: Own elaboration.} \label{fig2}
\end{figure*}

The island model is one of the most widespread parallel architectures \cite{cantu_survey, harada_survey}. In this paradigm (see Figure \ref{fig2}), the total population is divided into several semi-isolated subpopulations (demes) that evolve independently on separate processors \cite{cantu_survey, alba_survey, londe_biased_2025, whitley_island_1999}. These islands periodically interact through a process called migration, where copies of high-quality individuals are exchanged according to a predefined connection topology \cite{harada_survey, cantu_survey, londe_biased_2025, da_silveira_parallel_2019}. This isolation allows different demes to follow unique search trajectories through the solution space, maintaining higher genetic diversity and preventing the entire population from being dominated by a single local optimum \cite{alba_survey, whitley_island_1999}. Migration acts as a trigger for rapid evolutionary change, drawing a direct parallel to the biological theory of punctuated equilibria, which posits that evolutionary progress often occurs in bursts following a period of stability \cite{cantu_survey, skolicki}.

The structure of the migration graph determines three fundamental graph-theoretic properties that govern information flow across the system: the \textit{diameter} --- the longest shortest path between any two islands, which bounds the minimum number of migration events a solution requires to reach every island --- the \textit{average path length}, which captures the mean number of hops between island pairs and largely determines the speed at which an elite solution
disseminates globally, and the \textit{clustering coefficient}, which measures the tendency of neighboring islands to also be mutually connected, preserving local co-adaptation structures that would be disrupted by overly aggressive global mixing~\cite{alba_impact_topology, gog_matching}. The interplay between these three properties defines the fundamental exploration-exploitation trade-off of the topology: graphs with large diameter and high clustering promote diversity and independent search trajectories, while low-diameter graphs accelerate convergence at the risk of premature population homogenization~\cite{lissovoi_impact, alba_behavior}.

Among static topologies, the \textit{ring} and \textit{fully-connected} graphs represent the two extremes of this spectrum. In a ring, each island communicates only with its immediate neighbors, yielding a diameter of $\lfloor N/2 \rfloor$ and slow but diverse information propagation that allows islands to accumulate distinct local adaptations over extended periods before exchanging solutions~\cite{dang_ring, lissovoi_sparse}. The fully-connected graph, by
contrast, makes every island a direct neighbor of every other, reducing the diameter to one and maximizing migration speed at the cost of rapid convergence toward a shared gene pool~\cite{lissovoi_impact, cantu_survey}. Classical intermediate structures such as the \textit{hypercube} and the \textit{mesh} offer $O(\log N)$ and $O(\sqrt{N})$ diameters respectively, and have been widely adopted as default topologies in distributed GA frameworks~\cite{cantu_survey, harada_survey}. More expressive topologies arise from complex network theory. \textit{Small-world networks}, introduced by Watts and Strogatz~\cite{watts_strogatz}, are constructed by rewiring a fraction $p$ of the edges of a ring lattice with random long-range connections. The resulting graphs simultaneously achieve high local clustering --- preserving neighborhood co-adaptation structures reminiscent of a ring --- and short average path lengths comparable to those of a random graph, dramatically accelerating the global dissemination of elite solutions without sacrificing local diversity~\cite{watts_strogatz, alba2026novel}. \textit{Scale-free networks}, generated via the preferential attachment mechanism of Barabási and Albert~\cite{barabasi_albert}, exhibit a power-law degree distribution in which a small number of highly connected hub islands emerge naturally, acting as super-connectors that relay elite individuals across the system in very few
migration hops while the majority of low-degree islands retain sufficient isolation to sustain independent evolutionary trajectories~\cite{barabasi_albert, gog_matching}. This structural heterogeneity mirrors the role of highly connected hub populations in biological metapopulation dynamics, where a few central demes disproportionately mediate gene flow between otherwise isolated groups.

Beyond static structures, recent work has explored \textit{dynamic topologies} that modify the migration graph during the run in response to the evolving state of the population~\cite{alba_survey, luong_multiagent, gong_dynamic}. \textit{Gossip-based} protocols, inspired by epidemic spreading models in distributed systems, assign each island a randomly resampled set of $f$ communication partners at every migration event, producing epidemic coverage that grows exponentially with the number of rounds and ensuring that no island becomes
permanently isolated regardless of the graph's instantaneous state~\cite{luong_multiagent, araujo_design}. \textit{Adaptive} topologies take a more principled approach: migration links are added or removed based on pairwise diversity metrics between subpopulations, pruning connections between islands that have converged to similar solutions --- where migration would inject redundant genetic material --- and establishing new bridges toward genetically
distant islands where it would yield the greatest informational benefit~\cite{luong_multiagent, gong_dynamic}.

Equally critical to topology is the definition of \textit{migration policies}, which govern three interdependent decisions: which individuals leave an island (\textit{emigration selection}), how many are transferred per event (\textit{migration size}), and how incoming individuals are absorbed into the receiving population (\textit{immigration replacement})~\cite{cantu_migration, araujo_design, alba_behavior}. Elite-based emigration --- selecting the highest-fitness individuals as migrants --- is the most widespread strategy, on
the grounds that they carry the most valuable genetic building blocks and are therefore most likely to benefit the receiving island~\cite{cantu_migration, fitness_brkga_paper}. Diversity-based selection strategies take a complementary perspective, prioritizing individuals that are genetically distant from the receiving island's current population regardless of their absolute fitness, thereby maximizing the informational novelty of each migration event~\cite{replacement_diversity, dm_limga}. On the receiving end, the most common replacement policy substitutes the worst-performing residents with incoming migrants, combining imported quality with local diversity~\cite{cantu_migration, araujo_design}, while diversity-aware policies instead target the most similar resident to the migrant in genotype space, preserving existing fitness levels while maximizing structural novelty~\cite{replacement_diversity, dm_limga}. Migration size modulates the intensity of each exchange: large transfers accelerate convergence but risk homogenizing the receiving population, while small transfers maintain diversity at the cost of slower information propagation~\cite{skolicki, araujo_design}. In
the context of BRKGA, elite-based emigration is the natural choice, as the biased crossover operator already identifies and preserves high-quality allele combinations within each island; exporting these elite individuals to neighboring islands amplifies the same mechanism at the inter-population level, enabling rapid propagation of superior loading configurations across the entire parallel system~\cite{fitness_brkga_paper, goncalves_parallel}.

Crucially, the optimal interval does not operate independently of the topology: a ring requires higher migration frequency to achieve the same global dissemination speed as a small-world or scale-free graph operating at a lower rate, and sparse topologies have been shown to be more robust to suboptimal interval choices precisely because their larger diameter provides a natural
buffer against premature convergence~\cite{lissovoi_sparse, dang_ring}. This interdependence between topology structure, migration parameters, and individual selection policies motivates a joint evaluation of all three dimensions, which this work carries out specifically in the context of the 3D Trailer Loading Problem.

Despite these advances, two gaps remain underexplored in the literature. First, existing parallel BRKGA studies rarely address the problem of fair cross-configuration comparison: when different parallelization strategies produce different numbers of solutions under the same wall-clock time, performance differences conflate solution count with solution quality, making conclusions unreliable. Second, while the influence of migration topology and individual
selection policies on island model behavior is well established theoretically, their specific impact in the context of a highly constrained combinatorial problem such as the 3D-TLP remains largely uncharacterized. This work addresses both gaps directly.

\section{METHODOLOGY OVERVIEW}

\subsection{BRKGA}

The proposed genetic algorithm is a Biased Random-Key Genetic Algorithm (BRKGA) based on the formulation presented by \cite{goncalves_biased_2013}. The overall evolutionary framework follows the principles of this approach, in which candidate solutions are represented by chromosomes composed of random keys and subsequently transformed into loading configurations through a problem-specific decoder. The main differences with respect to the formulation of \cite{goncalves_biased_2013} concern the chromosome decoding procedure and the fitness function, both of which have been adapted to the specific characteristics and operational constraints of the trailer loading problem considered in this study.

\subsubsection{Chromosome representation and decoding}

The chromosome incorporates the information required to construct a complete loading solution. As in \cite{goncalves_biased_2013}, genes are used to determine the loading sequence of the packages and their orientation. However, the proposed representation introduces two additional decision components associated with the placement of each package within the available space.

\begin{figure}[!ht]
\includegraphics[width=\linewidth]{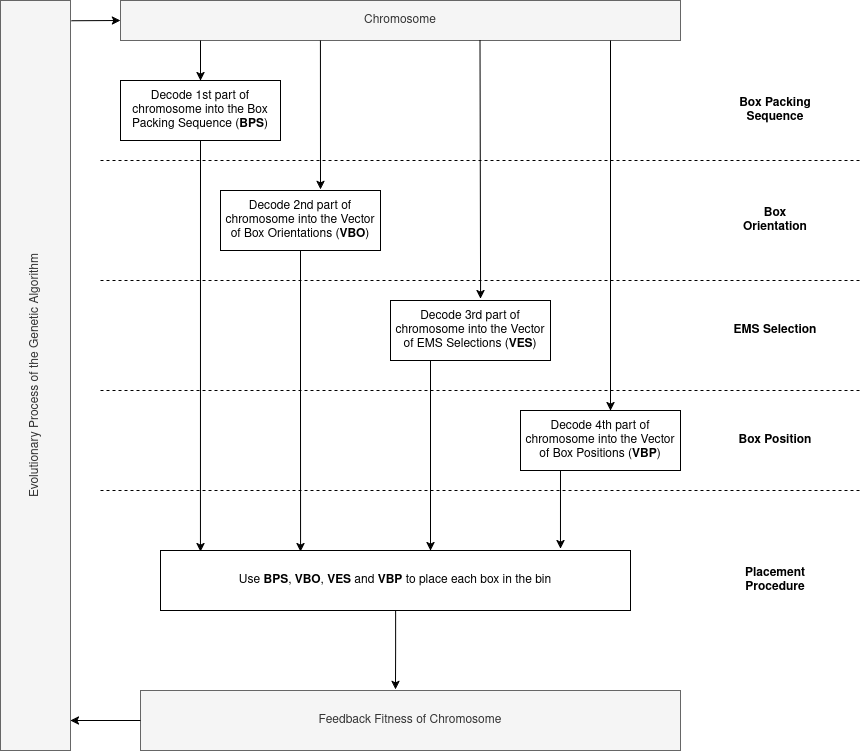}
\caption[Architecture of the algorithm.]{Architecture of the algorithm. Adapted from: \cite{goncalves_biased_2013}} \label{fig1}
\end{figure}

In particular, additional genes determine the Empty Maximal Space (EMS) selected for the placement of each package and the position of the package within the selected EMS. This represents a significant modification of the decoding strategy. In the approach of \cite{goncalves_biased_2013}, the EMS selection is guided by the DFTRC-2 heuristic, while the package is positioned at the corner corresponding to the smallest coordinates within the selected space. In the proposed approach, these deterministic decisions are partially replaced by information encoded directly in the chromosome, as can be seen in Figure \ref{fig1}.

The introduction of these additional genes increases the flexibility of the decoder because the algorithm is no longer restricted to a single deterministic choice of EMS and a fixed corner placement. Instead, different EMSs and different positions within each selected space can be explored during the evolutionary search. Consequently, packages can also be placed at positions that are not necessarily coincident with the lowest-coordinate corner of the available space, including more central positions within an EMS.

This modification expands the effective search space of the algorithm. Rather than relying on a predefined placement rule to determine a large part of the loading configuration, the evolutionary process can directly explore alternative combinations of loading order, package orientation, EMS selection, and position. This greater flexibility is particularly relevant for the problem considered, where the feasibility and quality of a loading pattern depend on several interacting geometric and logistical constraints, rather than simply maximizing the utilized volume or the loaded weight.

\subsubsection{Fitness function}

The second major modification concerns the fitness function used to evaluate the chromosomes. Unlike conventional BPP formulations, where the objective is often primarily related to volume or weight utilization, the proposed approach incorporates a broader set of criteria reflecting the physical and operational requirements of real trailer loading. In particular, feasibility depends not only on non-overlap and available capacity, but also on factors such as the support provided to stacked packages, cargo stability during transportation, weight distribution, package fragility, unloading sequence, and the physical feasibility of the resulting loading pattern.

Accordingly, the fitness evaluation considers the amount of cargo loaded subject to the available volume and weight capacity, together with 32 logistical and safety-related requirements detailed in Table \ref{tab:criteria}. These requirements include weight distribution across the loading platform, package blocking and securing, the positioning of fragile packages, the weight placed on top of other packages, and compliance with the required unloading sequence. The decoder therefore evaluates not only whether packages can be geometrically placed within the trailer, but also whether the resulting configuration satisfies the conditions required for a feasible and operationally appropriate load.

\begin{table*}[t]
\caption{Evaluation criteria}
\label{tab:criteria}
\centering
\begin{tabular}{lll}
\hline
\textbf{Criteria family} & \textbf{Description} & \textbf{Criteria amount} \\ \hline
Loaded value & Takes into account the amount of packages loaded & 1\\
Stability and locking & Check that the load is secured and stable to prevent it from shifting or falling & 11\\
Weight distribution & Assess the weight distribution on the platform & 11\\
Stacking & Analyse the stacking of the packages according to their characteristics & 7\\
Logistical & Evaluate logistical constraints & 3\\
\hline
\end{tabular}
\end{table*}

This formulation introduces a broader optimization objective in which cargo utilization is considered jointly with feasibility, safety, and operational requirements. Consequently, a solution with high space utilization is not necessarily preferred if it results in an unstable load, unsuitable weight distribution, or violations of other logistical constraints. The proposed fitness function thus guides the evolutionary search toward loading configurations that are both space-efficient and suitable for practical load transportation.

% -----------------------------------------------------------
\subsection{ Island-based parallelization framework}
% -----------------------------------------------------------

The modifications introduced in both the chromosome representation and the fitness function have an important computational consequence. On the one hand, adding genes for EMS selection and package positioning considerably enlarges the search space that must be explored by the evolutionary algorithm, and on the other hand, the more comprehensive fitness function
requires the evaluation of a larger number of geometric, logistical, and safety conditions for every decoded chromosome.

As a result, the computational effort required to decode and evaluate each individual is substantially greater than in a formulation based on a more restricted placement strategy and a simpler fitness evaluation. This becomes particularly relevant as the population size and number of generations increase, since a large number of candidate loading configurations must be constructed and assessed throughout the evolutionary process.

To address this computational cost, a novel island-based parallelization framework for genetic algorithms (PANGEA) was implemented and integrated with the modified BRKGA algorithm.
Instead of relying on a single population evolving sequentially, the search is distributed among multiple populations or islands that can explore different regions of the solution space in parallel. This architecture is particularly suitable for the proposed formulation because the expanded chromosome representation creates a large number of possible loading configurations, while the computationally demanding decoding and fitness evaluation can be distributed across the available islands.

The island-based implementation therefore serves two complementary purposes. First, it provides a mechanism for reducing the computational time associated with evaluating
the evolutionary population. Second, the existence of multiple evolving populations increases the potential diversity of the search, allowing different regions of the expanded solution space to be explored simultaneously. The parallelization is thus a consequence not only of the computational requirements introduced by the proposed decoder and fitness function, but also of the need to efficiently exploit the greater flexibility of the new BRKGA formulation.

The island model is governed by three groups of configurable parameters. First, the \emph{number of islands and processor allocation} determines how the available computational budget is divided among simultaneously evolving populations and how many cores each island can use for fitness evaluation. Second, the \emph{communication topology} specifies the graph structure through which islands exchange individuals, governing the speed and reach of solution propagation. Third, the \emph{migration parameters} , the interval and size, control the frequency and scale of these inter-island exchanges. Together, these three groups define the exploration-exploitation trade-off of the parallel search; each is described in turn below, while Table \ref{tab:params} presents an overview of them all.

% ----- Summary table -----

\begin{table*}[!t]
  \caption{Configurable parameters of the island-based parallelization framework (PANGEA).}
  \label{tab:params}
  \centering
  \renewcommand{\arraystretch}{1.25}
  \begin{tabular}{p{3.2cm} p{8.2cm} p{2.6cm} l}
    \hline
    \textbf{Parameter} & \textbf{Role} & \textbf{Values evaluated} & \textbf{Status} \\
    \hline
    \multicolumn{4}{l}{\textit{BRKGA algorithm}} \\
    \hline
    Population size per island
      & Number of individuals in each island's subpopulation. Determines
        genetic diversity within each island and the computational cost
        per generation.
      & 160 & Fixed \\
    Elite fraction ($\rho_e$)
      & Proportion of the population kept as elites each generation. Elites
        are preserved unchanged and act as parents with higher inheritance
        probability during crossover.
      & 0.20 & Fixed \\
    Mutant fraction ($\rho_m$)
      & Proportion of new individuals generated with fully random keys each
        generation, injecting fresh genetic material and preventing premature
        convergence.
      & 0.20 & Fixed \\
    Elite crossover bias ($\rho$)
      & Probability that each gene in a crossover offspring is inherited from
        the elite parent rather than the non-elite parent. Controls the
        exploitation pressure of the biased crossover operator.
      & 0.70 & Fixed \\
    \hline
    \multicolumn{4}{l}{\textit{Stopping criteria}} \\
    \hline
    Max generations
      & Hard upper bound on the number of evolutionary iterations per run,
        regardless of convergence state.
      & 2000 & Fixed \\
    Time limit
      & Wall-clock time limit per experiment. Ensures fair computational
        budget comparison across configurations independent of convergence
        speed.
      & 5400\,s (90\,min) & Fixed \\
    Early stopping (patience)
      & Maximum consecutive generations without global improvement before
        execution halts. Prevents unnecessary computation once convergence
        is detected.
      & 60 generations & Fixed \\
    \hline
    \multicolumn{4}{l}{\textit{Island-based parallelization}} \\
    \hline
    Number of islands
      & Degree of population decentralization. More islands promote diversity
        but reduce the per-island computational budget.
      & 1, 5, 10 & Varied \\
    Cores per island
      & Processor cores for parallel fitness evaluation within each island.
        Total budget (30 cores) is constant.
      & 30, 6, 3 & Varied \\
    Inner model
      & Parallelization strategy within each island for fitness evaluation.
        Master-slave distributes individual evaluations across the available
        cores of that island.
      & Master-slave & Fixed \\
    Nested parallelism
      & Enables two-level parallelism: island model at the outer level
        (population diversity) and master-slave at the inner level (fitness
        evaluation throughput).
      & True & Fixed \\
    CPU budget policy
      & Strategy for distributing available cores across islands when the
        total budget does not divide evenly.
      & Best-effort & Fixed \\
    \hline
    \multicolumn{4}{l}{\textit{Communication topology}} \\
    \hline
    Topology type
      & Graph structure determining which islands can exchange individuals.
        Governs solution propagation speed and inter-island genetic diversity.
      & Ring, \newline Fully Connected & Varied \\
    \hline
    \multicolumn{4}{l}{\textit{Migration}} \\
    \hline
    Migration interval
      & Generations between consecutive migration events. Shorter intervals
        accelerate elite dissemination; longer intervals preserve diversity.
      & 5, 15 & Varied \\
    Migration size
      & Number of individuals transferred per migration event. Larger sizes
        accelerate convergence but increase homogenization risk.
      & 1, 5 & Varied \\
    Migration policy
      & Strategy for selecting which individuals emigrate from an island.
        Rotating-FBMP cycles through a set of fitness-based migration
        policies across successive events, combining different selection
        pressures.
      & Rotating-FBMP & Fixed \\
    Emigration policy
      & Determines whether emigrants are removed from the source island or
        remain as copies. \emph{Remove} prevents elites from accumulating
        in their origin island after emigrating.
      & Remove & Fixed \\
    Acceptance policy
      & Rule governing how incoming migrants are accepted into the receiving
        island's population. Auto selects the policy based on population
        state.
      & Auto & Fixed \\
    Migrant deduplication
      & Prevents sending duplicate individuals across islands within a
        sliding window of recent migrants. Reduces redundant genetic
        material transfer and promotes diversity.
      & Enabled, window\,=\,64 & Fixed \\
    Migration transport
      & Backend mechanism for inter-island data transfer.
      & Local (shared memory) & Fixed \\
    \hline
    \multicolumn{4}{l}{%
      \footnotesize
      PANGEA additionally supports Watts-Strogatz small-world,
      Barabási-Albert scale-free, Gossip, and Adaptive topologies,
      not evaluated in this work.}
  \end{tabular}
\end{table*}

\subsubsection{Number of islands and processor allocation}

The island model requires specifying two closely related parameters: the number of islands and the number of processor cores allocated to each island for the parallel evaluation of the
fitness function. Together, these parameters determine how the total available computational budget is distributed across the simultaneously evolving populations.

The number of islands governs the degree of population decentralization. A single island concentrates the entire computational budget within one population, maximizing the
resources available for fitness evaluation but providing no opportunity for independent parallel exploration. As the number of islands increases, the search is distributed among a larger number of simultaneously evolving populations, each exploring a different region of the solution space. This promotes greater population diversity and reduces the risk of premature convergence, at the cost of reducing the computational resources --- and therefore the evaluation throughput --- available to each individual island.

The processor allocation per island is the complementary parameter. Given a fixed total core budget, assigning more cores to fewer islands increases the inner parallelism available for
fitness evaluation within each population, while assigning fewer cores to more islands shifts the balance towards maintaining a larger number of independent populations.

An important architectural feature of the implementation is that the parallelism operates at two nested levels. At the outer level, the island model distributes the evolutionary search across independent populations, each running on a dedicated set of processor cores. At the inner level, each island employs a master-slave strategy to distribute the fitness evaluation of its individuals across the cores assigned to it. In this scheme, a master process manages the evolutionary loop --- selection, crossover, and population bookkeeping --- while slave processes evaluate candidate loading configurations in parallel. The two levels are therefore complementary: the island model targets population diversity and the exploration of different regions of the solution space, whereas the inner master-slave targets the throughput of the computationally expensive decoding and fitness evaluation step.

This two-level structure makes the number of islands and per-island core allocation a genuine trade-off: concentrating cores in few islands maximizes evaluation throughput but forgoes
the diversity benefits of multiple populations, while distributing cores across many islands promotes independent search trajectories at the cost of reduced per-island evaluation
capacity. The fitness evaluation of the 3D-TLP --- which involves decoding the chromosome into a three-dimensional loading plan and assessing a broad set of geometric, logistical, and safety criteria --- is particularly demanding, making the inner master-slave parallelism a critical component for maintaining acceptable execution times regardless of the number of islands chosen.

A degenerate case of particular interest is the single-island configuration, in which all available cores are assigned to a single population. This configuration eliminates inter-island communication entirely, making topology and migration parameters irrelevant, and therefore serves as a natural reference point against which the benefits of multi-island parallelization can be assessed.

\subsubsection{Communication topologies}

In PANGEA, the communication topology defines the neighborhood relationships between islands and, consequently, which islands can exchange individuals during migration events. In this work, two topologies are considered: Ring and Fully Connected. These topologies represent two contrasting levels of inter-island connectivity and allow the effect of communication intensity on the evolutionary process to be evaluated.

\textbf{Ring topology:} In the Ring topology, the islands are arranged in a cyclic structure, with each island connected to its two immediate neighbors. For an island $i$ in a system of $N$ islands, its neighbors are $(i-1)\bmod N$ and $(i+1)\bmod N$. Therefore, each island has a fixed degree of two, and genetic material propagates progressively through the network. This restricted connectivity allows the subpopulations to evolve more independently, preserving differences between islands and limiting the rapid dissemination of individuals. Consequently, improvements discovered by one island may require several migration events to reach distant islands.

\textbf{Fully Connected topology:} In the Fully Connected topology, every island is directly connected to all other islands. Each island therefore has $N-1$ neighbors, and any individual selected for migration can be transmitted directly to any other island in a single migration event. This configuration promotes rapid information sharing and facilitates the dissemination of high-quality solutions throughout the population. However, the higher connectivity also increases the interaction between subpopulations and may lead to faster homogenization of the islands.

The two topologies therefore provide different balances between independent exploration and information sharing. The Ring topology favors a more gradual propagation of genetic material, whereas the Fully Connected topology promotes rapid dissemination across the entire population. Evaluating both configurations makes it possible to analyze the effect of inter-island connectivity on the convergence and solution quality of the proposed approach.

\textbf{Future topologies:} PANGEA is also being extended with additional, more adaptive communication mechanisms, including Gossip-based and Adaptive topologies. These approaches aim to dynamically modify the communication structure according to the evolutionary state of the islands, potentially providing a more flexible balance between exploration and exploitation. Their integration and evaluation in the context of the 3D-TLP are part of ongoing work and are outside the scope of the present study.

\subsubsection{Migration parameters} In addition to the topology, two parameters determine the characteristics of the migration process: the migration interval and the migration size. The migration interval defines the number of evolutionary generations between consecutive migration events. Shorter intervals result in more frequent communication between islands, allowing promising genetic material to spread more rapidly. Conversely, longer intervals allow the subpopulations to evolve independently for a greater number of generations before exchanging information.

The migration size defines the number of individuals transferred during each migration event. A smaller migration size limits the influence that one island can exert on another and helps preserve differences between subpopulations. Increasing the migration size facilitates the transfer of a larger amount of genetic material and can accelerate the dissemination of promising solutions, but may also increase the degree of similarity between islands.

The effects of the migration interval and migration size are closely related to the selected topology. For the same migration frequency and migration size, the Fully Connected topology allows genetic material to reach a larger number of islands directly, whereas propagation in the Ring topology occurs progressively through neighboring islands. The combination of topology and migration parameters therefore determines the balance between information exchange and population diversity. For this reason, the two topologies are evaluated jointly with different migration intervals and migration sizes in the experimental study.

% ============================================================
%  SECTION IV — VALIDATION AND RESULTS
% ============================================================

\section{VALIDATION AND RESULTS}

\subsection{Dataset}

The proposed algorithm is evaluated using a dataset derived from 50 real trailer transportation operations involving large-sized packages. The dataset contains 687 packages distributed across the 50 transport instances. The number of packages per trailer varies according to the characteristics of each load, particularly its weight and volume, resulting in instances with different sizes and levels of complexity.

The packages comprise 25 different orthohedral dimensional configurations. However, their distribution is not uniform. The most frequent configuration, 2850 × 2100 mm, accounts for 224 packages (32.6\%), followed by 2440 × 1220 mm, with 172 packages (25.0\%), and 3660 × 2100 mm, with 52 packages (7.6\%). These three configurations represent approximately 65.2\% of the entire dataset, while the remaining 34.8\% corresponds to the other 22 dimensional configurations. Although length (L) and width (W) show relatively concentrated distributions, package height (H) exhibits greater variability, which increases the difficulty of constructing compact three-dimensional loading patterns.

A relevant characteristic of the dataset is that package orientation is constrained by its physical handling requirements. Packages equipped with double-entry supports can be positioned using either (L × W × H) or (W × L × H), allowing the length and width to be exchanged while maintaining the same height. For packages without double-entry supports, this rotation is not feasible. Their original orientation must be preserved, with the longest dimension aligned with the longitudinal axis of the trailer.

The 50 real instances also provide sufficient variability to assess whether parallel populations can explore different regions of the solution space and obtain robust loading configurations across loads with different characteristics. Thus, the dataset enables the algorithm to be evaluated under conditions that closely reflect the constraints encountered in practical trailer loading operations, rather than under purely geometric packing assumptions.

% -----------------------------------------------------------
\subsection{Implementation details}
% -----------------------------------------------------------

The proposed approach is implemented in Python 3.12.3 within PANGEA, the framework developed as part of this work to support parallel and distributed evolutionary algorithms. The implementation uses NumPy, Pandas, and Matplotlib, in versions 1.26.4, 2.1.4, and 3.8.4, respectively, among other libraries. Python's \textit{multiprocessing} module is used to parallelize the evaluation of individuals within each island. The implementation was developed following the structure of an existing BRKGA implementation, which implements the BRKGA described in \cite{goncalves_biased_2013}, and was used as a reference for the organization of the evolutionary components\footnote{\textit{BRKGA repository: https://github.com/dasvision0212/3D-Bin-Packing-Problem-with-BRKGA}}. The resulting implementation was adapted to the specific requirements of the 3D-TLP and integrated into PANGEA to support the island-based architecture.

\subsection{Experimental design}

\subsubsection*{Island configurations}

Three island configurations are evaluated. The single-island configuration $1\times30$ allocates all 30 cores to one population and serves as a reference point: with only one island, inter-island communication does not occur and the topology and migration parameters are not applicable. The two multi-island configurations, $5\times6$ and $10\times3$, distribute the 30-core budget across 5 and 10 independent populations, respectively, allocating 6 and 3 cores per island for parallel fitness evaluation. In all experiments, ten populations are maintained overall; therefore, configurations with fewer than ten islands are executed independently and repeatedly until ten populations are covered ($5\times6$ is run twice; $10\times3$ covers the ten populations in a single run).

\subsubsection*{Communication topologies}

Two topologies are evaluated: Ring and Fully Connected, as described in Section III B.2. Their performance is compared to assess the effect of inter-island connectivity on convergence and solution quality.

\subsubsection*{Migration parameters}

Two migration parameters are varied in the experiments. The migration interval takes the values 5 and 15 generations, controlling how frequently islands exchange individuals. The migration size takes the values 1 and 5 individuals per event, controlling the volume of genetic material transferred at each exchange. The remaining migration settings are fixed: the
selection policy is Rotating-FBMP, which cycles through a set of fitness-based migration policies across successive events; the emigration policy removes migrants from the source island rather than retaining copies; the acceptance policy is set to auto; and
migrant deduplication is enabled with a sliding window of 64 recent migrants to prevent redundant transfers. Inter-island data transfer uses the local shared-memory transport.

\subsubsection*{BRKGA parameters}

The BRKGA parameters are fixed across all experiments. Each island maintains a population of 160 individuals. The elite fraction is $\rho_e = 0.20$, the mutant fraction $\rho_m = 0.20$, and the elite crossover bias $\rho = 0.70$. The inner parallelization model is master-slave with nested parallelism enabled and a best-effort CPU budget policy.

\subsubsection*{Stopping criteria}

Each run is subject to three stopping criteria, whichever is reached first: a maximum of 2000 generations, a wall-clock time limit of 5400\,s (90\,min), and an early-stopping patience of 60 consecutive generations without improvement in the global best fitness. The time limit ensures a fair computational budget across all island configurations regardless of per-generation cost.

Crossing the island configurations with the topology and migration parameters yields 17 GA configurations in total. The single-island baseline ($1\times30$) contributes one configuration, as topology and migration do not apply. Each of the two multi-island configurations ($5\times6$ and $10\times3$) is combined with 2 topologies, 2 migration intervals, and 2 migration sizes, producing $2\times2\times2 = 8$ configurations per island type. The total is therefore $1 + 8 + 8 = 17$ configurations.

\subsection{Results}

\subsubsection{Representativeness of the fitness function}

To assess how accurately the fitness function reflects solution quality, the proportions of feasible and infeasible solutions were compared across fitness ranking positions for each GA configuration. A solution is considered \textit{feasible} when the resulting loading plan satisfies the logistical and safety requirements of the transport operation. Feasibility is assessed by an expert who qualitatively evaluates the generated loading plan based on practical criteria and professional experience. Conversely, a solution is considered \textit{infeasible} when the expert identifies one or more logistical or safety conditions that would prevent the loading plan from being practically acceptable. Thus, feasibility reflects not only the geometric validity of the loading configuration but also its suitability for real-world operation.

\begin{figure}[!ht]
\includegraphics[width=\linewidth]{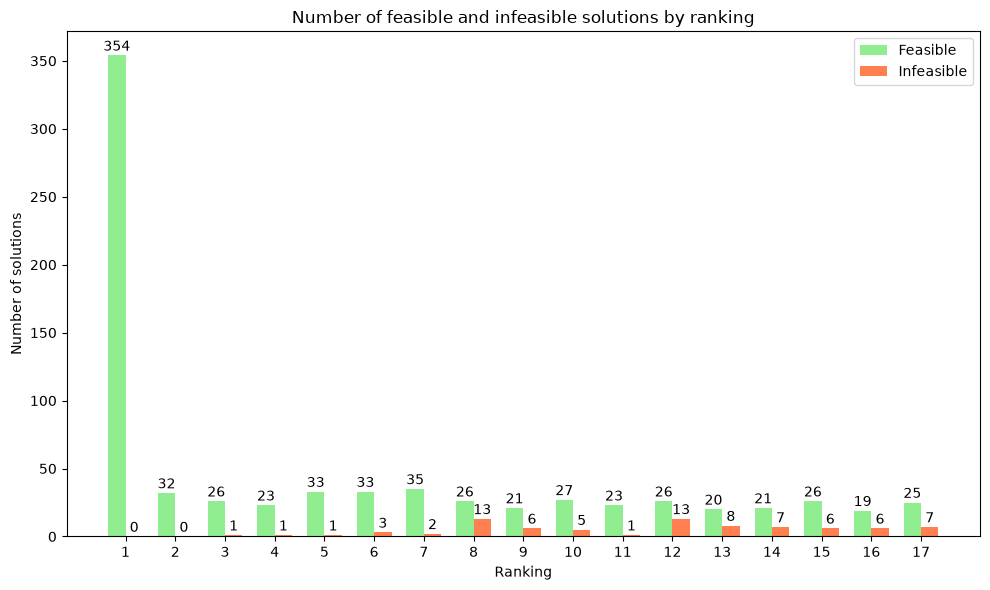}
\caption[Feasibility by fitness ranking.]{Feasibility by fitness ranking.} \label{fig3}
\end{figure}

As shown in Figure \ref{fig3}, solutions with better fitness values (i.e., those at the top of the ranking) tend to be feasible, whereas infeasible solutions are predominantly found among those with poorer fitness scores (i.e., lower-ranked solutions).

\subsubsection{Transport complexity}

\begin{figure}[!ht]
\includegraphics[width=\linewidth]{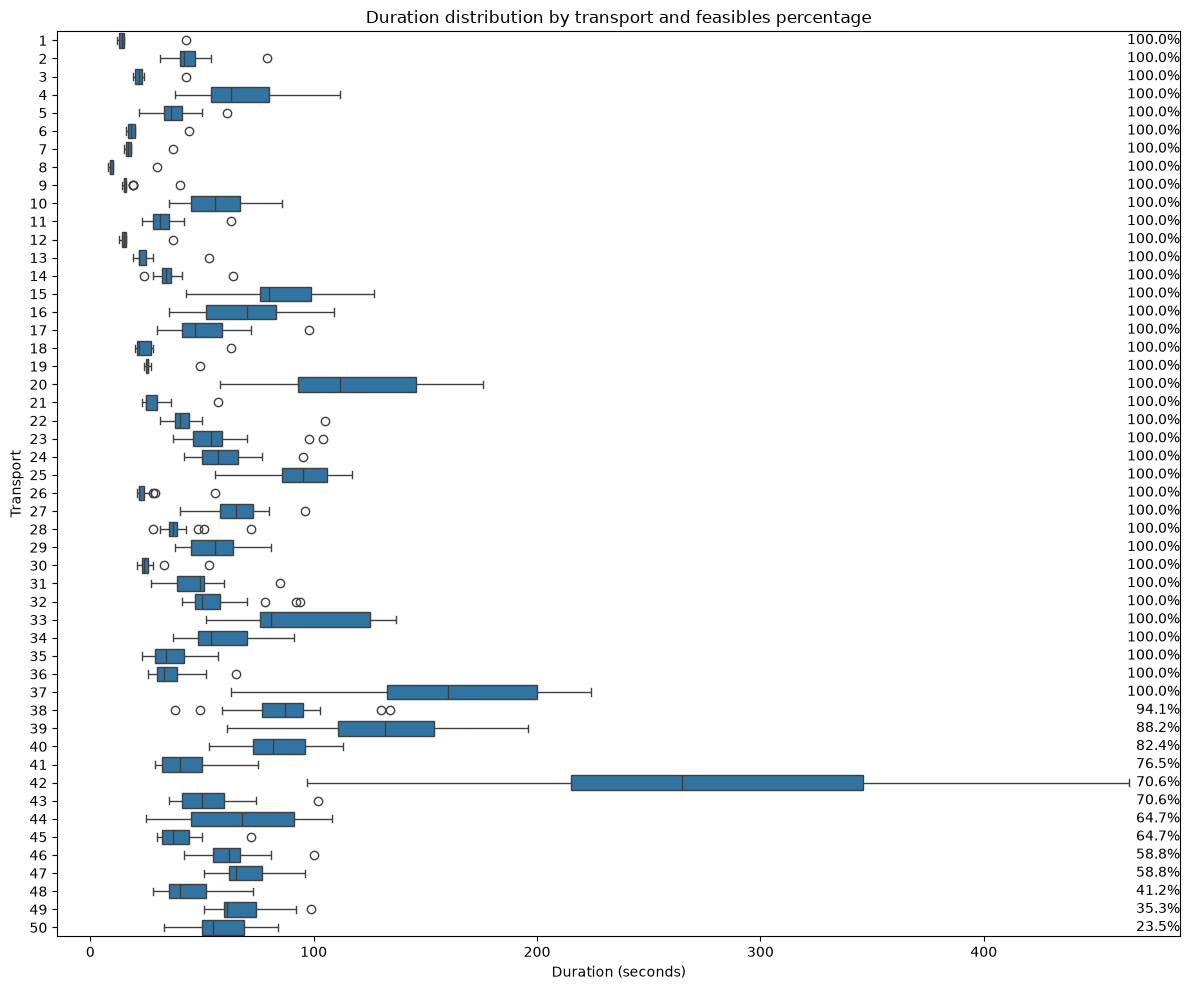}
\caption[Duration distribution by transport and feasibles percentage.]{Duration distribution by transport and feasibles percentage.} \label{fig4}
\end{figure}

The dataset comprises transport instances of varying difficulty. To illustrate this, Figure \ref{fig4} presents the distribution of computational times across the different transport instances for the 17 GA configurations considered. The right-hand margin reports the percentage of feasible solutions obtained for each transport instance.

For 37 transport instances (74\%), all GA configurations obtain a feasible solution within 200 seconds. For the remaining 13 instances, feasible solutions are obtained, but not consistently across all configurations. In the most challenging case, feasible solutions are obtained by only 4 out of the 17 configurations. Thus, feasible solutions are found for all transport instances in at least 4 configurations.

One transport instance stands out due to its substantially higher computational time compared with the others. Nevertheless, its execution time remains within reasonable limits (less than 10 minutes) for the practical use of the proposed tool.

\begin{figure}[!ht]
\includegraphics[width=\linewidth]{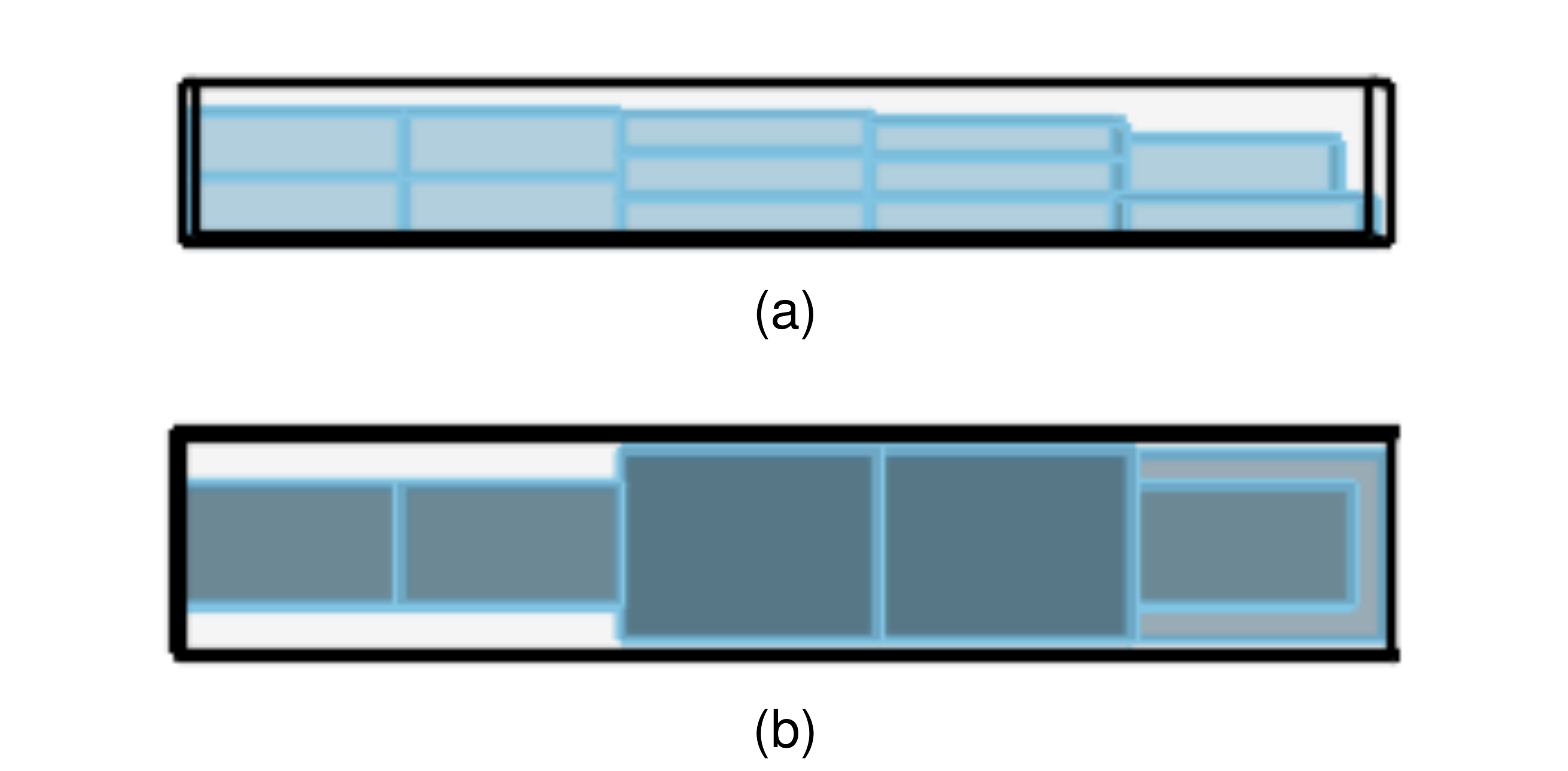}
\caption[Lateral (a) and aerial (b) view of a feasible solution of transport instance 6. Truck cab on the left-hand side.]{Lateral (a) and aerial (b) view of a feasible solution of transport instance 6. Truck cab on the left-hand side.} \label{fig7}
\end{figure}

\begin{figure}[!ht]
\includegraphics[width=\linewidth]{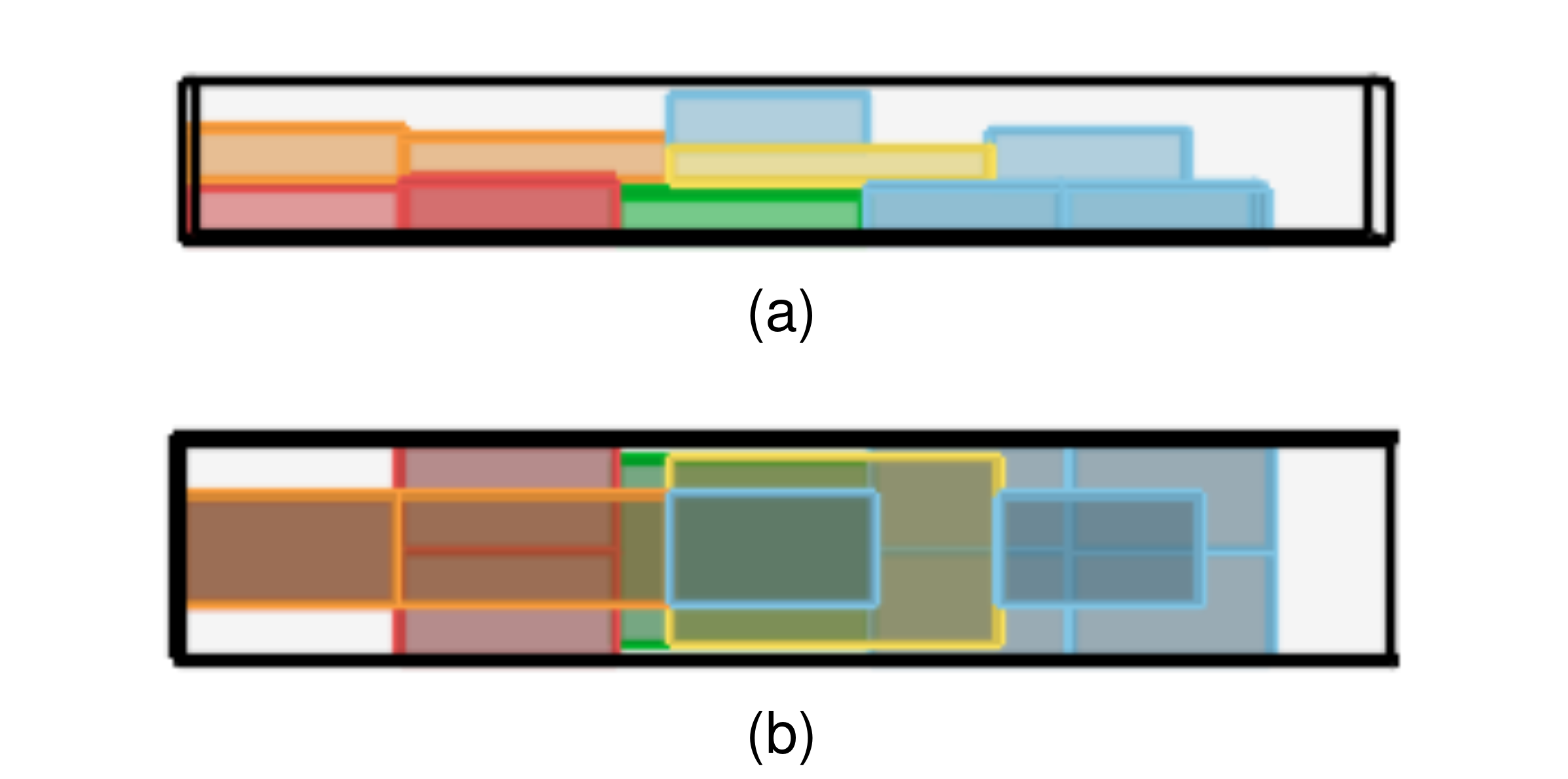}
\caption[Lateral (a) and aerial (b) view of a feasible solution of transport instance 49. Truck cab on the left-hand side.]{Lateral (a) and aerial (b) view of a feasible solution of transport instance 49. Truck cab on the left-hand side.} \label{fig8}
\end{figure}

To illustrate the variability in difficulty across the transport instances, Figures \ref{fig7} and \ref{fig8} present the lateral and aerial view of two examples of feasible loading solutions generated by the algorithm. Figure \ref{fig7} shows a feasible solution for transport instance 6 in Figure \ref{fig4} with 12 packages of different sizes from the same customer, for which all 17 configurations obtain a feasible solution, resulting in a 100\% feasibility rate. Moreover, the execution time remains below 21 seconds for all parallel configurations, while the non-parallelized baseline requires approximately 44 seconds.

In contrast, Figure \ref{fig8} presents a feasible solution obtained for instance 49 in Figure \ref{fig4}, which is among the most challenging instances in the dataset. In this case, involving 13 packages of different sizes from 5 different customers (unloading sequence by color: blue, yellow, orange, green, and red), feasible solutions are obtained by only 35.3\% of the configurations. The increased difficulty of this instance can be attributed to several factors, including the variability in package dimensions and the number of distinct customers that must be considered during unloading.

In general, the difficulty of an instance increases with the number of loading and logistical requirements that must be satisfied simultaneously. In particular, the unloading sequence and stackability requirements may introduce additional conflicts, as the dimensions of the packages can make it difficult or even impossible to satisfy both criteria at the same time. Consequently, instances involving a larger number of criteria from Table \ref{tab:criteria} tend to pose greater challenges to the algorithm, since a feasible solution must simultaneously satisfy multiple, and potentially conflicting, constraints.

\subsubsection{Genetic Algorithms performance}

\begin{figure}[!ht]
\includegraphics[width=\linewidth]{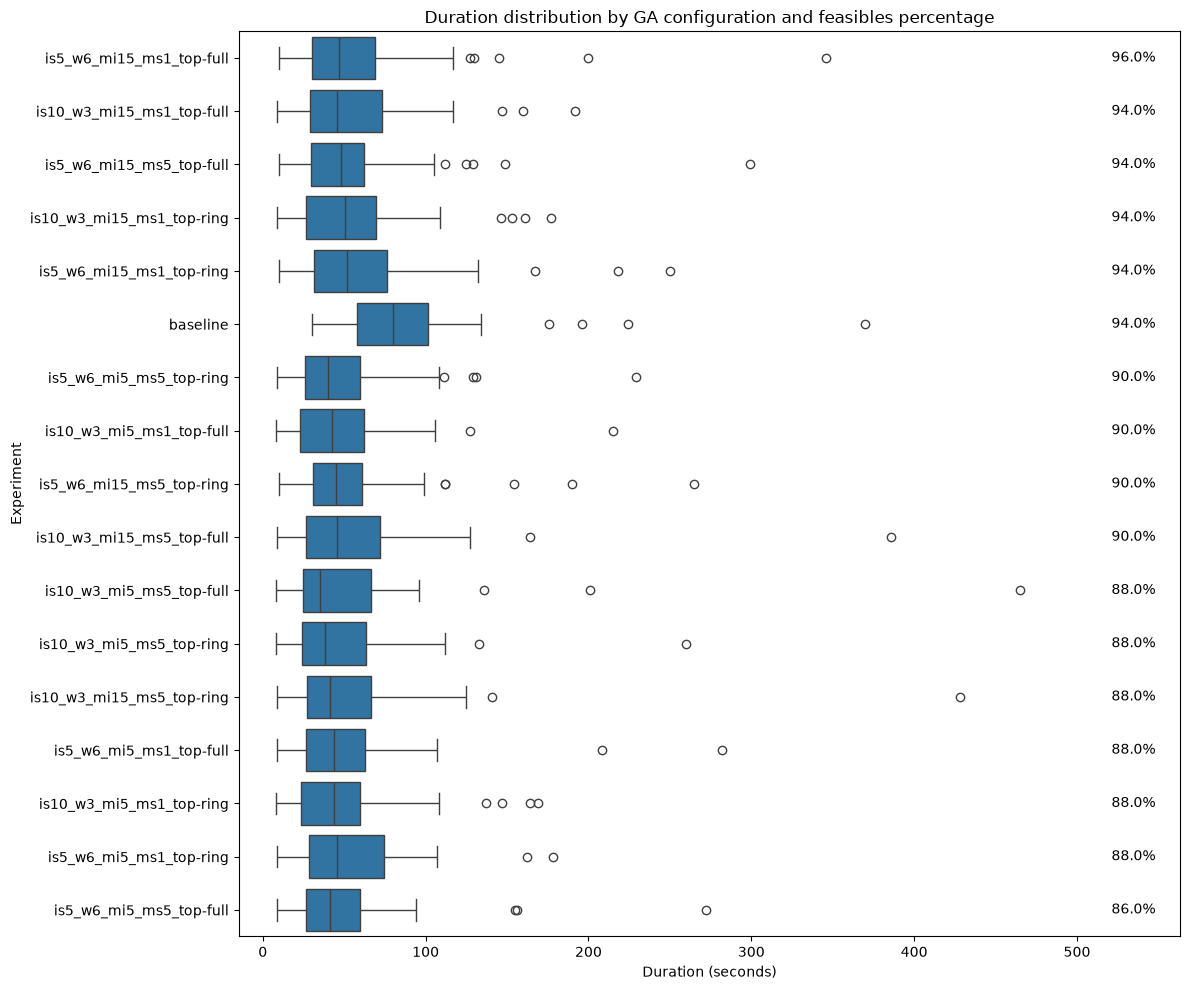}
\caption[Duration distribution by GA configuration and feasibles percentage.]{Duration distribution by GA configuration and feasibles percentage.} \label{fig5}
\end{figure}

In relation to the performance of the 17 different GA configurations, Figure \ref{fig5} shows the distribution of computational time and the percentage of feasible solutions obtained by each configuration. The configurations are ordered from top to bottom according to their percentage of feasible solutions, with ties being resolved based on the median execution time. Execution times are similar across configurations, except for the single-island configuration (baseline), which, as expected, requires more computational time than the others.

Regarding the feasibility rate of the different GA configurations, only minor differences can be observed, with all configurations achieving feasible solutions for at least 86\% of the transport instances. The configuration achieving the highest feasibility rate employs a fully connected topology, 5 islands, 6 cores per island, a migration interval of 15, and a migration size of 1. This configuration obtains feasible solutions for 48 out of the 50 instances analyzed.

The two instances for which the best configuration has not found a viable solution are instances 48 and 50 in Figure \ref{fig4}. These represent the first and third-most challenging instances in the dataset, respectively, based on the percentage of feasible solutions obtained across the 17 configurations, which is 41.2\% for instance 48 and 23.5\% for instance 50. 

Despite the long execution time, the baseline configuration achieves one of the highest feasibility percentages (94\%).

\subsubsection{Genetic Algorithms parameters influence}

The convergence performance of the different values of GA parameters was also evaluated. In general, no substantial differences were observed among the different values, although certain parameter settings consistently yielded slightly better results. Importantly, all parallel configurations achieved a better average performance than the non-parallelized baseline.

\begin{figure}[!ht]
\includegraphics[width=\linewidth]{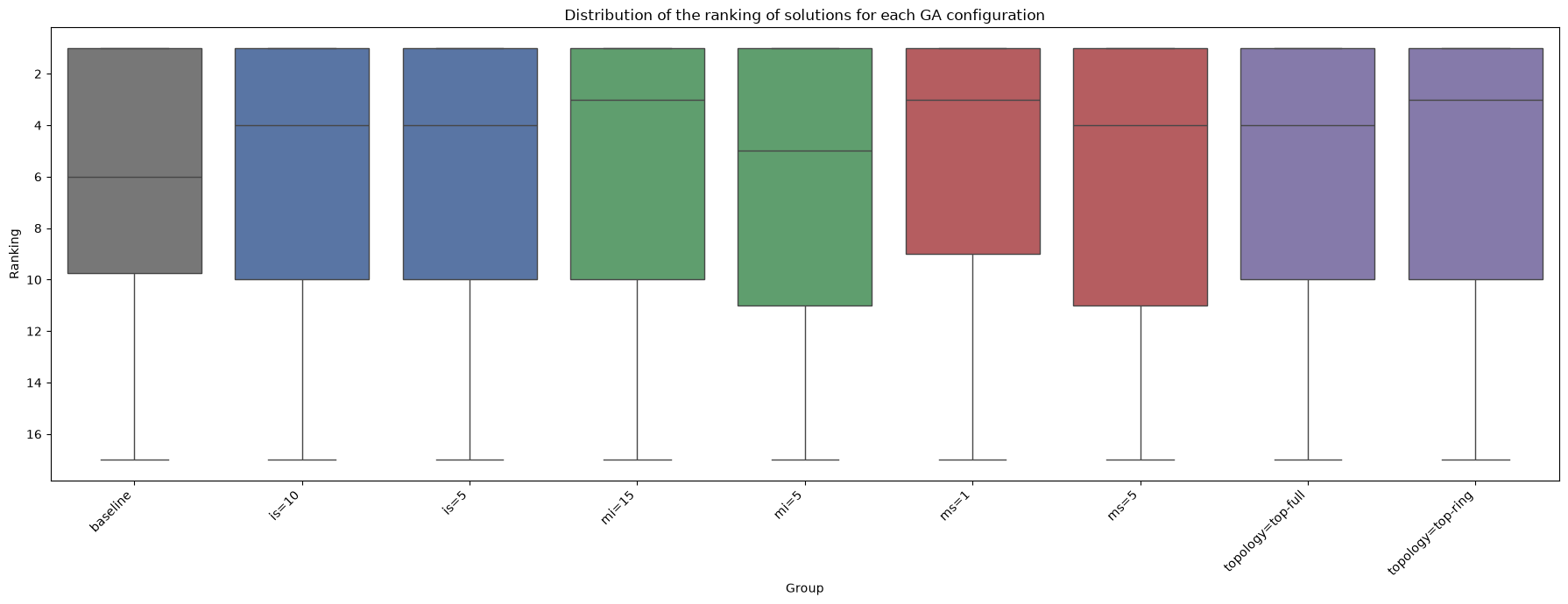}
\caption[Distribution of the ranking of solutions for each GA parameter value.]{Distribution of the ranking of solutions for each GA parameter value.} \label{fig6}
\end{figure}

Figure \ref{fig6} shows the distribution of the rankings obtained for the different parameter values. The distributions are broadly similar across parameter values, which can be partly attributed to the fact that, for many transport instances, multiple configurations obtain the same best solution and therefore share the first position in the ranking. The most noticeable differences are observed for the migration interval and migration size parameters, represented by the green and red boxplots, respectively. In these cases, the median ranking indicates better performance for a migration interval of 15 ($mi=15$) and a migration size of 1 ($ms=1$).

These observations are further supported by the ranking statistics for the different parameter values reported in Table \ref{tab:GAranking}, where \textit{n} denotes the number of transport instances evaluated, \textit{$\bar{x}$} the ranking sample mean, \textit{s} the ranking sample standard deviation, \textit{$Q_{1}$} the first quartile (25th percentile), \textit{Me} the median, \textit{$Q_{3}$} the third quartile (75th percentile), \textit{$P_{90}$} the 90th percentile, and \textit{$r_{1}\%$} the percentage of solutions ranked first. The baseline configuration exhibits the poorest overall performance, with a mean ranking of 6.42, a median of 6.0, and 34\% of solutions achieving the first position in the ranking. In contrast, parallel configurations consistently achieve better average results.

\begin{table}[!t]
\caption{Statistics of the ranking of solutions for each GA parameter value}
\label{tab:GAranking}
\centering
\begin{tabular}{llllllllll}
\hline
\textbf{GA} & \textbf{$n$} & \textbf{$\bar{x}$} & \textbf{$s$} & \textbf{$Q_{1}$} & \textbf{$Me$} & \textbf{$Q_3$} & \textbf{$P_{90}$} & \textbf{$r_{1}\%$} \\ \hline
Baseline & 50 & 6.42 & 5.17 & 1.0 & 6.0 & 9.75 & 15.0 & 34.00\\
top-full & 400 & 5.98 & 5.35 & 1.0 & 4.0 & 10.0 & 14.0 & 41.25\\
top-ring & 400 & 5.78 & 5.49 & 1.0 & 3.0 & 10.0 & 15.0 & 43.00\\
ms-1 & 400 & 5.55 & 5.20 & 1.0 & 3.0 & 9.0 & 14.0 & 43.75\\
ms-5 & 400 & 6.22 & 5.62 & 1.0 & 4.0 & 11.0 & 15.0 & 40.50\\
mi-5 & 400 & 6.22 & 5.64 & 1.0 & 5.0 & 11.0 & 15.0 & 41.25\\
mi-15 & 400 & 5.55 & 5.17 & 1.0 & 3.0 & 10.0 & 14.0 & 43.00\\
is-5 & 400 & 5.87 & 5.40 & 1.0 & 4.0 & 10.0 & 15.0 & 41.75\\
is-10 & 400 & 5.90 & 5.45 & 1.0 & 4.0 & 10.0 & 15.0 & 42.50\\
\hline
\end{tabular}
\end{table}

The differences between the two topologies are relatively small. Nevertheless, the ring topology achieves a better mean ranking and a higher proportion of first-ranked solutions than the fully connected topology. A more pronounced difference is observed for the migration parameters. Consistent with the distributions shown in Figure \ref{fig6}, a migration size of 1 outperforms a migration size of 5, while a migration interval of 15 yields better results than an interval of 5. In these cases, the difference in mean ranking is more noticeable than for the other parameters.

Finally, the number of islands has the smallest effect on convergence performance in this experiment. The different island configurations produce very similar results, with only marginal differences in both the mean ranking and the proportion of first-ranked solutions.

\section{CONCLUSIONS}
Results demonstrate that the proposed island-based parallel BRKGA is able to find feasible loading plans for all of the transport instances considered, while maintaining computational times that are compatible with the practical use of the tool. Beyond the overall feasibility rate, the experiments provide insight into how the characteristics of the transport instances and the different GA parameters affect the search process.

The analysis of the fitness function indicates that it provides a meaningful representation of solution quality. Solutions with better fitness values are predominantly feasible, whereas infeasible solutions are more frequently found among lower-ranked individuals. This relationship suggests that the fitness function effectively guides the search towards feasible regions of the solution space, which is particularly relevant for a problem characterized by multiple and potentially conflicting loading and logistical constraints.

The experiments also highlight the heterogeneous difficulty of the transport instances. While 74\% of the instances were solved feasibly by all configurations within 200 seconds, the remaining instances proved substantially more challenging. In the most difficult cases, feasible solutions were obtained by only a small subset of the configurations. The examples presented illustrate that this difficulty is not solely related to computational scale, but also to the interaction between different logistical requirements. In particular, unloading sequence and stackability constraints can conflict as a consequence of package dimensions, making some combinations of requirements considerably more restrictive. This suggests that the difficulty of an instance is strongly influenced by the number and interaction of constraints that must be satisfied simultaneously.

Regarding the GA configurations, the results show that the parallel approach consistently improves upon the non-parallelized baseline in terms of computational time and average convergence performance. The single-island configuration requires substantially more computational time, supporting the expected benefit of distributing the search among multiple islands. At the same time, the differences in feasibility rates among the configurations are relatively small: all configurations obtain feasible solutions for at least 86\% of the transport instances. This indicates that the proposed parallelization scheme is relatively robust with respect to the specific parameter combinations considered. Despite the long execution time, the baseline configuration achieves one of the highest feasibility percentages (94\%).

The convergence analysis provides further insight into the influence of the GA parameters. Although the performance differences between parameter values are generally moderate, the migration parameters appear to have a greater influence than the number of islands or the migration topology. In particular, a migration size of 1 and a migration interval of 15 are associated with better average rankings and a higher proportion of first-ranked solutions. This behavior may indicate that relatively small and moderately frequent exchanges of information provide a suitable balance between information sharing and the preservation of population diversity. However, these results should be interpreted within the scope of the experimental setting, rather than as evidence of a universally optimal parameter configuration.

In contrast, the number of islands has only a marginal effect on convergence performance in the experiments conducted. The configurations with 5 and 10 islands produce very similar ranking distributions, suggesting that increasing the number of islands beyond five does not provide a clear additional benefit under the computational resources and problem instances considered. Similarly, the differences between the ring and fully connected topologies are limited, although the ring topology obtains slightly better ranking statistics in this experiment.

The approach achieves a high feasibility rate across a heterogeneous set of instances, while keeping computational times within reasonable limits for practical application. At the same time, the results reveal that the most challenging instances are strongly influenced by the interaction between multiple logistical constraints, rather than by the GA configuration alone. Addressing these highly constrained cases therefore represents an important opportunity for further improvement.

Future work could focus on improving the treatment of highly constrained instances, particularly those in which unloading sequence and stackability requirements interact strongly. Finally, extending the experimental evaluation to larger datasets and additional computational environments would provide further evidence regarding the scalability and generalization of the proposed approach.

\section{ACKNOWLEDGEMENT}
This work was supported by the grant CPP2022-009530 (SMARTLOGISTIC project) funded by
the Ministry of Science and Innovation, through the AEI, within the framework of the State Program to Promote Scientific-Technical Research and its Transfer—part of the State Plan for Scientific, Technical and Innovation Research 2021-2023—and in the context of the Recovery, Transformation and Resilience Plan funded by Next Generation funds.

% References
\bibliographystyle{unsrt} % We choose the "unsrt" reference style (sorted by order of appearance)
\bibliography{biblio}

\end{document}